\documentclass[runningheads,a4paper]{llncs}

\usepackage{amssymb}
\usepackage{amsmath}
\usepackage{graphicx}
\usepackage{multirow}
\usepackage[normalem]{ulem}
\useunder{\uline}{\ul}{}
\usepackage{bbding}
\usepackage{ifxetex}
\usepackage{ifluatex}

\ifxetex
  \usepackage[ios,font=seguiemj.ttf]{emoji}
  \usepackage{fontspec}
  
\else
  \ifluatex
    \usepackage[ios,font=Symbola_hint.ttf]{emoji}
    \usepackage{fontspec}
    
  \else
    \usepackage[T1]{fontenc}
    \usepackage[utf8]{inputenc}
    \usepackage[ios]{emoji}
  \fi
\fi
\usepackage{url}
\urldef{\mailsa}\path|{sanjaya, lakshika, amit, derek}@knoesis.org|
\newcommand{\keywords}[1]{\par\addvspace\baselineskip
\noindent\keywordname\enspace\ignorespaces#1}

\begin{document}

\mainmatter  % start of an individual contribution

\pdfinfo{
   /Author (Sanjaya Wijeratne, Lakshika Balasuriya, Amit Sheth, Derek Doran)
   /Title (EmojiNet: Building a Machine Readable Sense Inventory for Emoji)
   /Keywords (EmojiNet; Emoji Analysis; Emoji Sense Disambiguation)
   /Subject (EmojiNet: Building a Machine Readable Sense Inventory for Emoji - SocInfo 2016)
}

% first the title is needed
\title{EmojiNet: Building a Machine Readable Sense Inventory for Emoji}

% a short form should be given in case it is too long for the running head
\titlerunning{EmojiNet: Building a Machine Readable Sense Inventory for Emoji}

\author{Sanjaya Wijeratne \and Lakshika Balasuriya \and Amit Sheth \and Derek Doran}
\authorrunning{Wijeratne et al.}
% (feature abused for this document to repeat the title also on left hand pages)

\institute{Kno.e.sis Center, Wright State University\\
Dayton, Ohio, USA\\
\mailsa\\
\url{http://www.knoesis.org}}

%\toctitle{Lecture Notes in Computer Science}
%\tocauthor{Authors' Instructions}
\maketitle

\begin{abstract}
%Mimimum 70, Maximum 150

Emoji are a contemporary and extremely popular way to enhance electronic communication. Without rigid semantics attached to them, emoji symbols take on different meanings based on the context of a message. Thus, like the word sense disambiguation task in natural language processing, machines also need to disambiguate the meaning or `sense' of an emoji. In a first step toward achieving this goal, this paper presents EmojiNet, the first machine readable sense inventory for emoji. EmojiNet is a resource enabling systems to link emoji with their context-specific meaning. It is automatically constructed by integrating multiple emoji resources with BabelNet, which is the most comprehensive multilingual sense inventory available to date. The paper discusses its construction, evaluates the automatic resource creation process, and presents a use case where EmojiNet disambiguates emoji usage in tweets. EmojiNet is available online for use at \url{http://emojinet.knoesis.org}.


\end{abstract}

\section{Introduction}

Pictographs commonly referred to as `emoji' have grown from their introduction in the late 1990's by Japanese cell phone manufacturers to an incredibly popular form of computer mediated communication (CMC).  Instagram reported that as of April 2015, 40\% of all messages posted on Instagram consist of emoji~\cite{instagramstudy}. From a 1\% random sample of all tweets published from July 2013 to July 2016, the service Emojitracker reported its processing of over 15.6 billion tweets with emoji\footnote{\url{http://www.emojitracker.com/api/stats}}. Creators of the SwiftKey Keyboard for mobile devices also report that 6 billion messages per day contain emoji~\cite{swiftkeymost}. Even authorities on language use have acknowledged emoji; the American Dialect Society selected `eggplant' \emoji{1F346} as the most notable emoji of the year\footnote{\url{http://www.americandialect.org/2015-word-of-the-year-is-singular-they}}, and The Oxford Dictionary recently awarded `face with tears of joy' \emoji{1F602} as the word of the year in 2015\footnote{\url{http://blog.oxforddictionaries.com/2015/11/word-of-the-year-2015-emoji/}}. All these reports suggest that emoji are now an undeniable part of the world's electronic communication vernacular.

\begin{figure}
\centering
\includegraphics[width=1.0\linewidth]{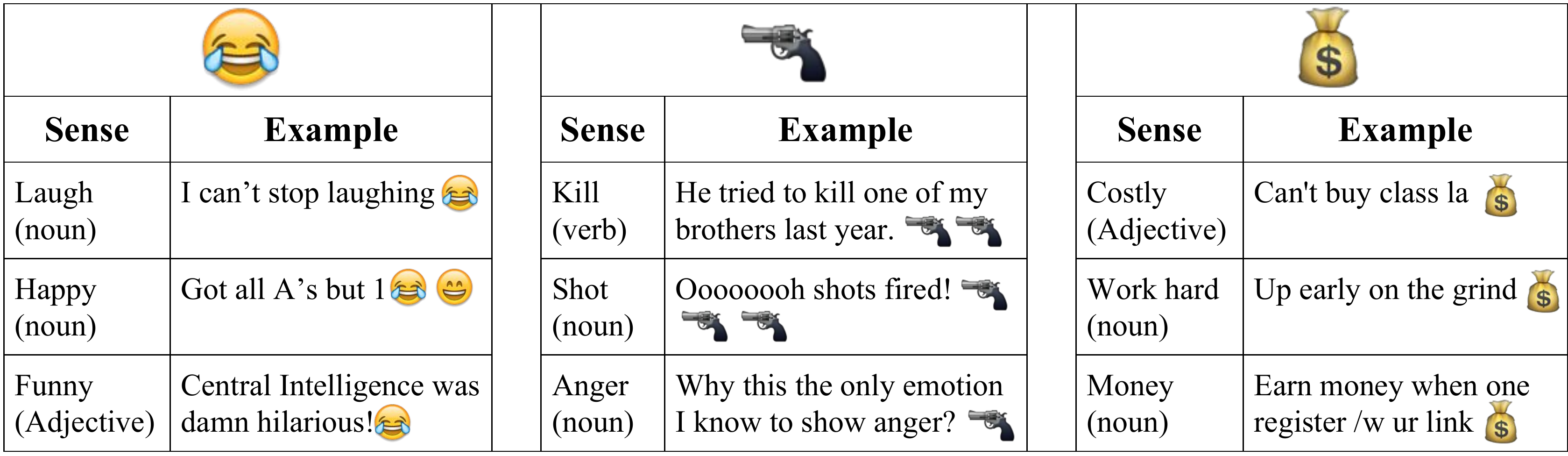} 
\caption{Emoji Usage in Social Media with Multiple Senses.}
\label{emojiensesexamples}
\end{figure}

People use emoji to add color and whimsiness to their messages~\cite{kelly2015characterising} and to articulate hard to describe emotions~\cite{emogireport}. Perhaps by design, emoji were defined with no rigid semantics attached to them\footnote{\url{http://www.unicode.org/faq/emoji_dingbats.html#4.0.1}}, allowing people to develop their own use and interpretation. Thus, similar to words, emoji can take on different meanings depending on context and part-of-speech~\cite{miller2016blissfully}. For example, consider the three emoji \emoji{1F602}, \emoji{1F52B}, and \emoji{1F4B0} and their use in multiple tweets in Figure~\ref{emojiensesexamples}. Depending on context, we see that each of these emoji can take on wildly different meanings. People use the \emoji{1F602} emoji to mean laughter, happiness, and humor; the \emoji{1F52B} emoji to discuss killings, shootings or anger; and the \emoji{1F4B0} emoji to express that something is expensive, working hard to earn money or simply to refer to money.

Knowing the meaning of an emoji can significantly enhance applications that study, analyze, and summarize electronic communications. For example, rather than stripping away emoji in a preprocessing step, sentiment analysis application reported in~\cite{novak2015sentiment} uses emoji to improve its sentiment score. However, knowing the meaning of an emoji could further improve the sentiment score. A good example for this scenario would be the \emoji{1F602} emoji, where people use it to describe both happiness (using senses such as laugh, joy) and sadness (using senses such as cry, tear). Knowing under which sense the emoji is being used could help to understand its sentiment better. But to enable this, a system needs to understand the particular meaning or {\em sense} of the emoji in a particular instance. However, no resources have been made available for this task~\cite{miller2016blissfully}. This calls for the need of a machine readable {\em sense inventory for emoji} that can provide information such as: (i) the plausible part-of-speech tags (PoS tags) for a particular use of emoji; (ii) the definition of an emoji and the senses it is used in; (iii) example uses of emoji for each sense; and (iv) links of emoji senses to other inventories or knowledge bases such as BabelNet or Wikipedia. Current research on emoji analysis has been limited to emoji-based sentiment analysis~\cite{novak2015sentiment}, emoji-based emotion analysis~\cite{wang2012harnessing}, and Twitter profile classification~\cite{lakshikagang,gangwordembeddings} etc. However, we believe introduction of an emoji sense inventory can open up new research directions on emoji sense disambiguation, emoji similarity analysis, and emoji understanding.

This paper introduces {\bf EmojiNet}, the first machine readable sense inventory for emoji. EmojiNet links emoji represented as Unicode with their English language meanings extracted from the Web. To achieve this linkage, EmojiNet integrates multiple emoji lexicographic resources found on the Web along with BabelNet~\cite{NavigliPonzetto:12aij}, a comprehensive machine readable sense inventory for words, to infer sense definitions. Our contributions in this work are threefold:

\begin{enumerate}
    \item We integrate four openly available emoji resources into a single, query-able dictionary of emoji definitions and interpretations;
    \item We use word sense disambiguation techniques to assign senses to emoji;
    \item We integrate the disambiguated senses in an open Web resource, EmojiNet, which is presently available for systems to query.
\end{enumerate}
The paper also discusses the architecture and construction of EmojiNet and presents an evaluation of the process to populate its sense inventory.

This paper is organized as follows. Section~\ref{sec:rr} discusses the related literature and frames how this work differs from and furthers existing research. Section \ref{sec:dc} discusses our approach and explains the techniques we use to integrate different resources to build EmojiNet. Section \ref{sec:eval} reports on the evaluation of the proposed approach and the evaluation results in detail. Section \ref{sec:con} offers concluding remarks and plans for future work.

\section{Related Work}\label{sec:rr}

Emoji was first introduced in the late 1990s but did not become a Unicode standard until 2009~\cite{unicodeimoji}. Following standardization, emoji usage experienced major growth in 2011 when the Apple iPhone added an emoji keyboard to iOS, and again in 2013 when the Android mobile platform began emoji support~\cite{instagramstudy}. In an experiment conducted using 1.6 million tweets, Novak {\em{et al.}} report that 4\% of them contained at least one emoji~\cite{novak2015sentiment}. Their recent popularity explains why research about their use is not as extensive as the research conducted on emoticons~\cite{miller2016blissfully}, which are the predecessor to emoji~\cite{novak2015sentiment} that used to represent facial expression, emotion or to mimic nonverbal cues in verbal speech~\cite{rezabek1998visual} in CMC.

Early research on emoji focuses on understanding its role in computer-aided textual communications. From interviews of 20 participants in close personal relationships, Kelly {\em{et al.}} reported that people use emoji to maintain conversational connections in a playful manner~\cite{kelly2015characterising}. Pavalanathan {\em{et al.}} studied how emoji compete with emoticons to communicate paralinguistic content on social media~\cite{pavalanathan2015emoticons}. They report that emoji were gaining popularity while emoticons were declining in Twitter communications. Miller {\em{et al.}} studied whether different emoji renderings would give rise to diverse interpretations~\cite{miller2016blissfully}, finding disagreements based on the rendering. This finding underscores the need for tools to help machines disambiguate the meaning and interpretation of emoji.

The Emoji Dictionary\footnote{\url{http://emojidictionary.emojifoundation.com/home.php?learn}} is a promising Web resource for emoji sense disambiguation. It is a crowdsourced emoji dictionary that provides emoji definitions with user defined sense labels, which are \texttt{word(PoS tag)} pairs such as \texttt{laugh(noun)}. However, it cannot be utilized by a machine for several reasons. First, it does not list the Unicode or short code names for emoji, which are common ways to programmatically identify emoji characters in text. Secondly, it does not list sense definitions and example sentences along with different sense labels for emoji. Typically, when using machine readable dictionaries, machines use such sense definitions and example sentences to generate contextually relevant words for each sense in the dictionary. Thirdly, the reliability of the sense labels is unclear as no validation of the sense labels submitted by the crowd is performed. With EmojiNet, we address these limitations by linking The Emoji Dictionary with other rich emoji resources found on the Web. This allows sense labels to be linked with their Unicode and short code name representations and discards human-entered sense labels for emoji that are not agreed upon by the resources. EmojiNet also links sense labels with BabelNet to provide definitions and example usages for different senses of an emoji.

\begin{figure}
\centering
\includegraphics[width=1.0\linewidth]{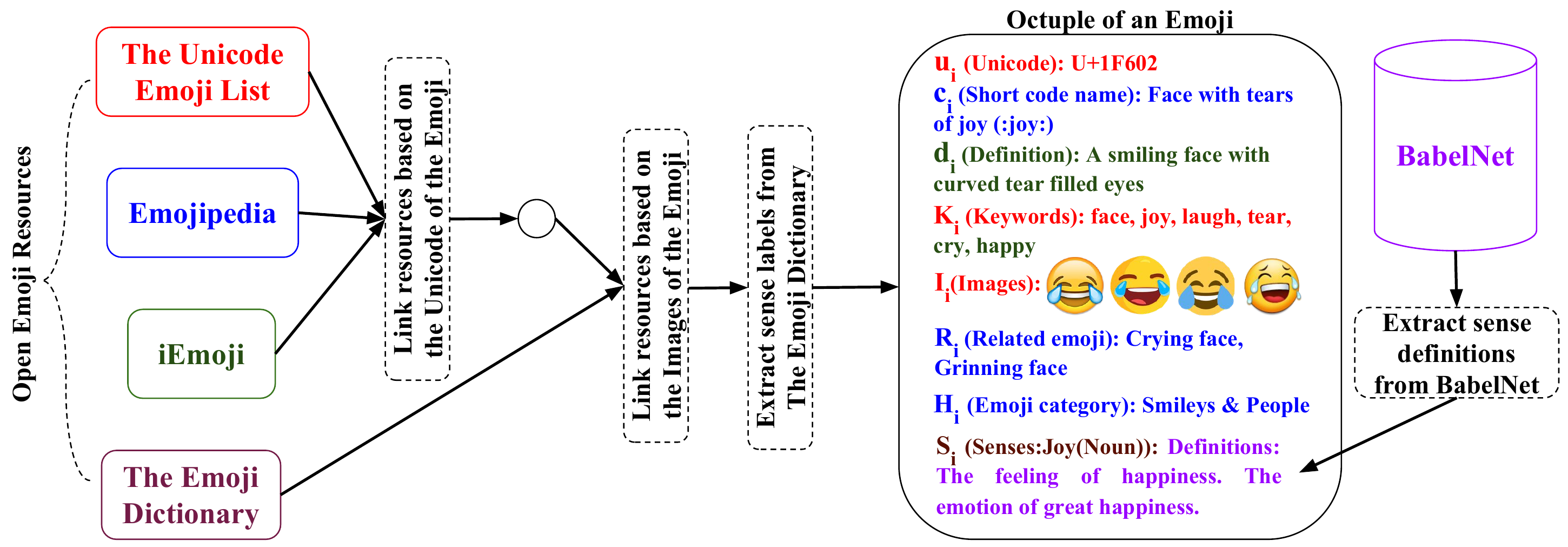} 
\caption{Bulding EmojiNet by Integrating Multiple Open Resources.}
\label{emojinet}
\end{figure}

\section{Building EmojiNet} \label{sec:dc}

We formally define EmojiNet as a collection of octuples representing the senses of an emoji. Let $E$ be the set of all emoji in EmojiNet. For each $e_i \in E$, EmojiNet records the octuple $e_i = (u_i, c_i, d_i, K_i, I_i, R_i, H_i, S_i)$, where $u_i$ is the Unicode representation of $e_i$, $c_i$ is the short code name of $e_i$, $d_i$ is a description of $e_i$, $K_i$ is the set of keywords that describe basic meanings attached to $e_i$, $I_i$ is the set of images that are used in different rendering platforms such as the iPhone and Android, $R_i$ is the set of related emoji for $e_i$, $H_i$ is the set of categories that $e_i$ belongs to, and $S_i$ is the set of different senses in which $e_i$ can be used within a sentence. An example for an octuple notation is shown as part of Figure~\ref{emojinet}. Each element in the octuple provides essential information for sense disambiguation. EmojiNet uses unicode $u_i$ and short code name $c_i$ of an emoji $e_i \in E$ to uniquely identify $e_i$, and hence, to search EmojiNet. $d_i$ is needed to understand what is represented by the emoji. It can also help to understand how an emoji should be used. $K_i$ is essential to understand different human-verified senses that an emoji could be used for. $I_i$ is needed to understand the rendering differences in each emoji based on different platforms. Images in $I_i$ can also help to conduct similar studies as~\cite{miller2016blissfully}, where the focus is to disambiguate the different representations of the same emoji on different rendering platforms. $R_i$ and $H_i$ could be helpful in tasks such as calculating emoji similarity and emoji sense disambiguation. Finally, $S_i$ is the key enabler of EmojiNet as a tool to support emoji sense disambiguation as $S_i$ holds all sense labels and their definitions for $e_i$ based on crowd and lexicographic knowledge. Next, we describe the open information EmojiNet extracts and integrates from the Web to construct the octuples.

\subsection{Open Resources}

Several emoji-related open resources are available on the Web, each carrying their own strengths and weaknesses. Some have overlapping information, but none has all of the elements required for a machine readable sense inventory. Thus, EmojiNet collects information across multiple open resources, linking them together to build the sense inventory. We describe the resources EmojiNet utilizes below. \\

\noindent
{\bf Unicode Consortium.} Unicode is a text encoding standard enforcing a uniform interpretation of text byte code by computers\footnote{\url{http://www.unicode.org/}}. The consortium maintains a complete list of the standardized Unicodes for each emoji\footnote{\url{http://www.unicode.org/emoji/charts/full-emoji-list.html}} along with manually curated keywords and images of emoji. Let the set of all emoji available in the Unicode emoji list be $E_U$. For each emoji $e_u \in E_U$, we extract the Unicode character $u_i$ of $e_u$, the set of all images $I_{e_{u}}$ associated with $e_u$ that are used to display $e_u$ on different platforms, and the set of keywords $K_{U_{e_{u}}} \subset K_{e_{u}}$ associated with $e_u$, where $K_{e_{u}}$ is the set of all manually-assigned keywords available for the emoji $e_u$. \\

\noindent
{\bf Emojipedia.} Emojipedia is a human-created emoji reference site\footnote{\url{https://en.wikipedia.org/wiki/Emojipedia}}. It provides Unicode representations for emoji, images for each emoji based on different rendering platforms, short code names, and other emoji manually-asserted to be related. Emojipedia organizes emoji into a pre-defined set of categories based on how similar the concepts are represented by each emoji, i.e., Smileys \& People, Animals \& Nature, or Food \& Drink. Let the set of all emoji available in Emojipedia be $E_P$. For each emoji $e_p \in E_P$, we extract the Unicode representation $u_{p}$, the short code name $c_p$, and the emoji definition $d_p$ of $e_{p}$, the set of related emoji $R_{e_{p}}$, and its category set $H_{e_{p}}$. \\

\noindent 
{\bf iEmoji.} iEmoji\footnote{\url{http://www.iemoji.com/}} is a service tailored toward understanding how emoji are being used in social media posts. For each emoji, it provides a human-generated description, its Unicode character representation, short code name, images across platforms, keywords describing the emoji, its category within a manually-built hierarchy, and examples of its use in social media (Twitter) posts. Let the set of all emoji available in iEmoji be $E_{IE}$. For each emoji $e_{ie} \in E_{IE}$, we collect the Unicode representation $u_{ie}$ of $e_{ie}$ and the set of keywords $K_{IE_{e_{ie}}} \subset K_{e_{ie}}$ associated with $e_{ie}$, where $K_{e_{ie}}$ is the set of all keywords available for $e_{ie}$.\\

\noindent
{\bf The Emoji Dictionary.} The Emoji Dictionary\footnote{\url{http://emojidictionary.emojifoundation.com/home.php?learn}} is a crowdsourced site providing emoji definitions with sense labels based on how they could be used in sentences. It organizes meanings for emoji under three part-of-speech tags, namely, nouns, verbs, and adjectives. It also lists an image of the emoji and its definition with example uses spanning multiples sense labels. Let the set of all emoji available in The Emoji Dictionary be $E_{ED}$. For each emoji $e_{ed} \in E_{ED}$, we extract its image $i_{e_{ed}} \in I_{ED}$, where $I_{ED}$ is the set of all images of all emoji in $E_{ED}$ and the set of crowd-generated sense labels $S_{e_{ed}}$.\\

\noindent
{\bf BabelNet}
BabelNet is the most comprehensive multilingual machine readable semantic network available to date~\cite{NavigliPonzetto:12aij}. It is a dictionary with lexicographic and encyclopedic coverage of words tied to a semantic network that connects concepts in Wikipedia to the words in the dictionary. It is built automatically by merging lexicographic data in WordNet with the corresponding encyclopedic knowledge extracted from Wikipedia\footnote{\url{http://babelnet.org/about}}. BabelNet has been shown effective in many research areas including word sense disambiguation~\cite{NavigliPonzetto:12aij}, semantic similarity, and sense clustering~\cite{camacho2015nasari}. For the set of all sense labels $S_{e_{ed}}$ in each $e_{ed} \in E_{ED}$, we extract the sense definitions and examples (if available) for each sense label $s_{e_{ed}} \in S_{e_{ed}}$ from BabelNet.

%\vspace{-15px}
\begin{table}[]
\centering
\caption{Emoji Data Available in Open Resources}
%\vspace{-5px}
\begin{tabular}{lr c c c c c c c c }
\multicolumn{1}{ c }{\textbf{Emoji Resource}} & \textbf{\textbf{$u$}} & \textbf{\textbf{$c$}} & \textbf{\textbf{$d$}} & \textbf{\textbf{$K$}} & \textbf{\textbf{$I$}} & \textbf{\textbf{$R$}} & \textbf{\textbf{$H$}} & \textbf{\textbf{$S$}} \\\hline 
Unicode Consortium                            & {\color{blue}\Checkmark}            & {\color{blue}\Checkmark}                & {\color{red}\XSolid}                     & {\color{blue}\Checkmark}                & {\color{blue}\Checkmark}            & {\color{red}\XSolid}                       & {\color{red}\XSolid}                      & {\color{red}\XSolid}                       \\ 
Emojipedia                                    & {\color{blue}\Checkmark}                      & {\color{blue}\Checkmark}                            &{\color{blue}\Checkmark}                  & {\color{red}\XSolid}                      & {\color{blue}\Checkmark}                            &{\color{blue}\Checkmark}                  & {\color{blue}\Checkmark}                  & {\color{red}\XSolid}                       \\ 
iEmoji                                        &{\color{blue}\Checkmark}                      & {\color{blue}\Checkmark}                            & {\color{blue}\Checkmark}                            & {\color{blue}\Checkmark}               & {\color{blue}\Checkmark}                            & {\color{red}\XSolid}                          & {\color{blue}\Checkmark}                         & {\color{red}\XSolid}                       \\ 
The Emoji Dictionary                          & {\color{red}\XSolid}                       & {\color{red}\XSolid}                       & {\color{red}\XSolid}                       & {\color{red}\XSolid}                      & {\color{blue}\Checkmark}             & {\color{red}\XSolid}                        & {\color{red}\XSolid}                    & {\color{blue}\Checkmark}                   \\ 
%BabelNet & ? & ? & ? & ? & ? & ? & ? & ? \\
\end{tabular}
\label{emojidataextraction}
\end{table}
%\vspace{-15px}

\noindent Table~\ref{emojidataextraction} summarizes the data about an emoji available across the four open resources. A `{\color{blue}\Checkmark}' denotes the availability of the information in the resource where `{\color{red}\XSolid}' denotes the non-availability. It is important to note that unique crowds of people deposit information about emoji into each resource, making it important to integrate the same type of data across many resources. For example, the set of keywords $K_i$, the set of related emoji $R_i$, and the set of categories $H_i$ for an emoji $e_i$ are defined by the crowds qualitatively, making it necessary to compare and scrutinize them to determine the elements that should be considered by EmojiNet. Data types that are `fixed', e.g. the Unicode $u_i$ of an emoji $e_i$, will also be useful to link data about the same emoji across the resources. We also note that The Emoji Dictionary uniquely holds the sense labels of an emoji, yet does not store its Unicode $u_i$. This requires EmojiNet to link to this resource via emoji images, as we discuss further in the next section.

\subsection{Integrating Emoji Resources} \label{imagealgo}
We now describe how EmojiNet integrates the open resources described above. The integration, illustrated in Figure~\ref{emojinet}, starts with the Unicode's emoji characters list as it is the official list of 1,791 emoji accepted by the Unicode Consortium for support in standardized software products. Using Unicode character representation in the emoji list, we link these emoji along with the information extracted from Emojipedia and the iEmoji websites. Specifically, for each emoji $e_u \in E_U$, we compare $u_u$, with all Unicode representations of the emoji in $E_P$ and $E_{IE}$. If there is an emoji $e_u \in E_U$ such that $u_u = u_p = u_{ie}$, we merge the three corresponding emoji $e_u \in E_U$, $e_p \in E_P$, and $e_{ie} \in E_{IE}$ under a single emoji representation $e_i \in E$. In other words, we merge all emoji where they share the same Unicode representation. We store all the information extracted from the merged resources under each emoji $e_i$ as the octuple described in Section~\ref{sec:dc}.

\subsubsection{Linking to The Emoji Dictionary} \label{emojiintegration}
Unfortunately, The Emoji dictionary does not store the Unicode of an emoji. Thus, we merge this resource into EmojiNet by considering emoji {\em images}. We created an index of multiple images of the 1,791 Unicode defined emoji in the Unicode Consortium website. We have downloaded a total of 13,387 images for the 1,791 emoji. These images are referred to as our example image dataset $I_{x}$. We additionally downloaded images of all emoji listed on The Emoji Dictionary website, which resulted in a total of 1,074 images. We refer to this set of images as the test image dataset $I_{t}$.

To align the two datasets, we implement a nearest neighborhood-based image matching algorithm~\cite{santos2010java} that matches each image in $I_{t}$ with the images in $I_{x}$. Because images are of different resolutions, we normalize them into  a 300x300px space and then divide them along a lattice of 25 non-overlapping regions of size 25x25px. We then find an average color intensity of each region by averaging its $R$, $G$ and $B$ pixel color values. To calculate the dissimilarity between two images, we sum the $L_2$ distance of the average color intensities of the corresponding regions. The final accumulated value that we receive for a pair of images will be a measure of the dissimilarity of the two images. For each image in $I_t$, the least dissimilar image from $I_x$ is chosen and the corresponding emoji octuple information is merged.

\begin{figure}
\centering
\includegraphics[width=1.0\linewidth]{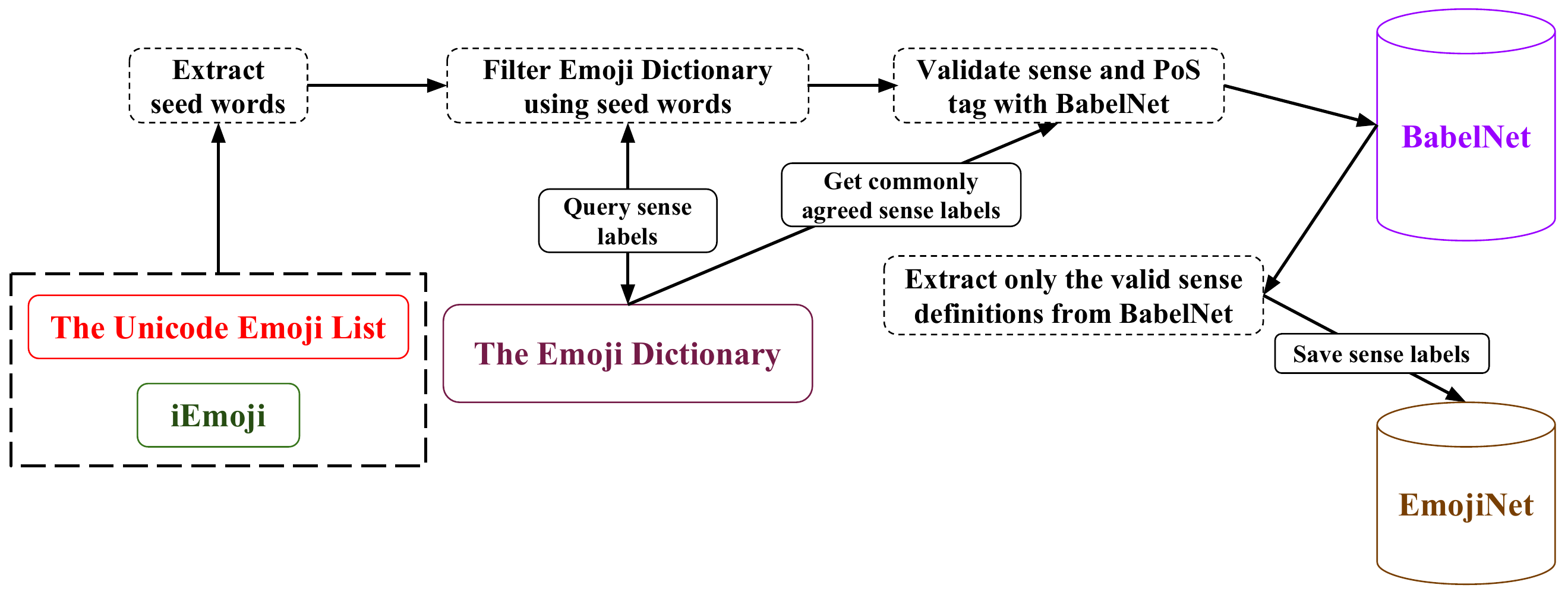} 
\caption{Emoji Sense and Part-of-Speech Filtering.}
\label{sensefiltering}
\end{figure}

\subsubsection{Emoji sense and part-of-speech filtering}
With The Emoji Dictionary linked to the rest of the open resources via images, EmojiNet can now integrate its sense and part-of-speech information (sense labels). However, as mentioned in Section \ref{sec:rr}, The Emoji Dictionary does not validate the sense labels collected from the crowd. Thus, EmojiNet must pre-process the sense labels from The Emoji Dictionary to verify its reliability. This is done in a three step process and it is elaborated in~Figure \ref{sensefiltering}. First, we use the set of keywords $K_i$ of emoji $e_i$ collected from the Unicode Consortium and iEmoji as seed words to identify reliable sense labels. These keywords are human-generated and represent the meanings in which an emoji can be used. For example, the \emoji{1F602} emoji has been tagged with the keywords {\em face, joy, laugh, and tear} in the Unicode emoji list and {\em tear, cry, joy, and happy} in the iEmoji website. Taking the union of these lists as a set of seed words, we filter the crowdsourced sense labels of an emoji from The Emoji Dictionary. For each keyword $k_i \in K_i$, we extract crowdsourced sense labels. For example, for the emoji \emoji{1F602}, The Emoji Dictionary lists three sense labels for the sense {\em laugh} as \texttt{laugh(noun)}, \texttt{laugh(verb)} and \texttt{laugh(adjective)}. However, the word {\em laugh} cannot be used as an adjective in the English language. Therefore, in the second step, we cross-check if the sense labels extracted from The Emoji Dictionary are valid using the information available in BabelNet. In this step, BabelNet reveals that {\em laugh} cannot be used as an adjective, so we discard \texttt{laugh(adjective)} and use \texttt{laugh(noun)}, \texttt{laugh(verb)} in EmojiNet. We do this for all seed keywords we obtain from the Unicode emoji list and iEmoji websites. In the final step, for any sense label in The Emoji Dictionary that is not a seed word but was submitted by more than one human (commonly agreed senses in Figure~\ref{sensefiltering}), EmojiNet validates these sense labels using BableNet. For example, the sense label \texttt{funny(adjective)} has been added by more than one user to The Emoji Dictionary as a possible sense for \emoji{1F602} emoji. This was not in our seed set; however, since there is common agreement on the sense label \texttt{funny(adjective)} and the word {\em funny} can be used as an adjective in the English language, EmojiNet extracts \texttt{funny(adjective)} from The Emoji Dictionary and adds it to its sense inventory under \emoji{1F602} emoji. Note that EmojiNet does not assign BabelNet sense IDs (or sense definitions) to the extracted sense labels yet. That process will require a word sense disambiguation step, which we will discuss in the next section.

\begin{figure}
\centering
\includegraphics[width=1.0\linewidth]{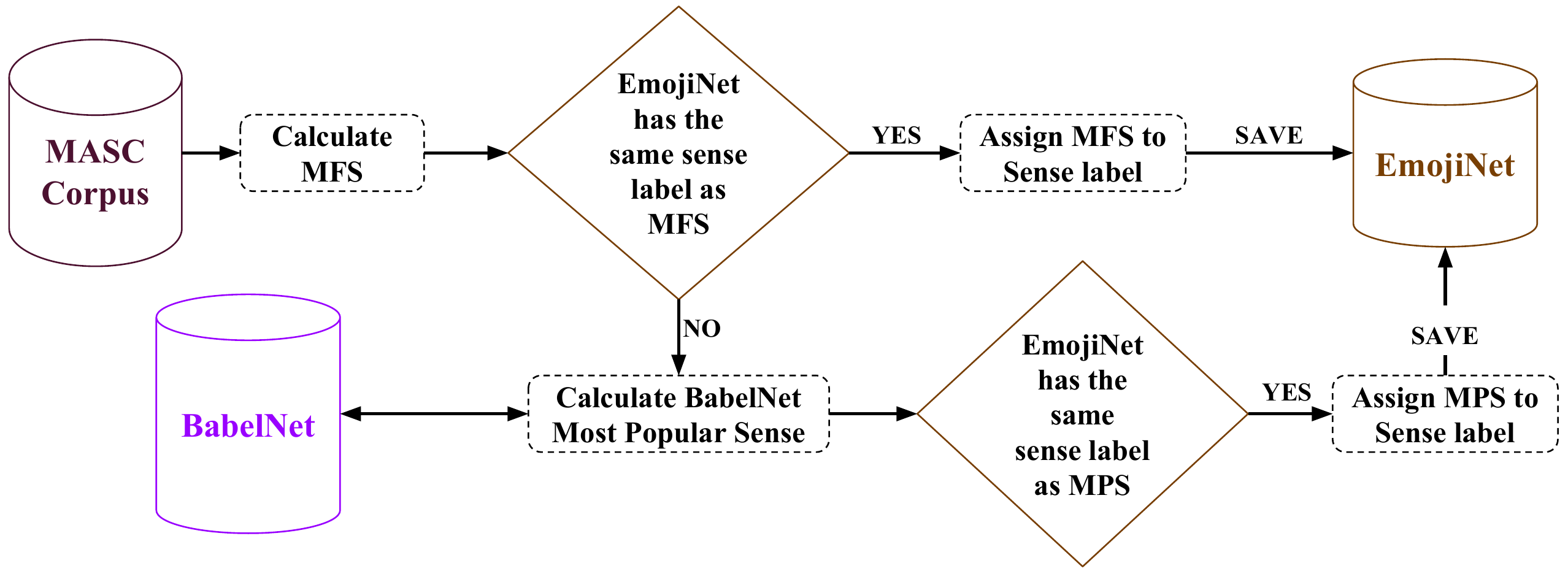} 
\caption{Using BabelNet to Assign Sense Definitions.}
\label{wsdimg}
\end{figure}

\subsection{Linking Emoji Resources with BabelNet} 
Having access to sense labels extracted from The Emoji Dictionary for each emoji, EmojiNet can now link these sense labels with BabelNet. This linking allows EmojiNet to interpret each sense label on how it can be used in a sentence. For example, the current version of BabelNet lists 6 different sense definitions for the sense label \texttt{laugh(noun)}. Thus, EmojiNet must select the most appropriate sense definition out of the six. As we described in Section~\ref{sec:rr}, The Emoji Dictionary does not link its sense labels with example sentences. Therefore, we cannot directly perform WSD on the sense labels or example sentences available in The Emoji Dictionary. Thus, to align the two resources, we use the MASC\footnote{\url{https://en.wikipedia.org/wiki/Manually_Annotated_Sub-Corpus_(MASC)}} corpus with a most frequent sense (MFS) baseline for WSD. MASC corpus is a balanced dataset that represents different text categories such as tweets, blogs, emails, letters, essays, and speech; words in the MASC corpus are already disambiguated using BabelNet~\cite{moro2014annotating}. We use these disambiguated words to calculate MFS for each word in the MASC corpus. Once the MFS of each word is calculated, for every sense label in EmojiNet, we assign its definition as the MFS of that same sense label retreived from MASC corpus. We use an MFS-based WSD baseline due to the fact that MFS is a very strong, hard-to-beat baseline model for WSD tasks~\cite{basile2014enhanced}. Figure~\ref{wsdimg} depicts the steps followed in our WSD approach.

EmojiNet has a total of 3,206 sense labels that need to be sense disambiguated using BabelNet. However, not all sense labels in EmojiNet were assigned BabelNet senses in the above WSD task. There were sense labels in EmojiNet which were not present in MASC corpus, hence they were not disambiguated. To disambiguate such sense labels which were left out, we define a second WSD task. We calculate the most popular sense (MPS) for each BabelNet sense, which we define as follows. For each BabelNet sense label $B_s$, we take the count of all sense definitions BabelNet lists for $B_s$. The MPS of $B_s$ is the BabelNet sense ID that has the highest number of definitions for $B_s$. If there are more than one MPS available, a sense ID will be picked randomly out of the set of MPS sense IDs as the MPS. Once the MPS is calculated, those will be assigned to their corresponding sense labels in EmojiNet which were left out in the first WSD task. Note that BabelNet holds multiple definitions that come from multiple resources such as WordNet, VerbNet, Wikipedia, etc. which are integrated into it. Hence, MPS of $B_s$ gives an indication of the popularity of $B_s$. With this step, we complete the integration of open resources to create EmojiNet.

\subsection{EmojiNet Web Application}
We expose EmojiNet as a web application at \url{http://emojinet.knoesis.org/}. The current version of EmojiNet supports searching for emoji based on Unicode character and short code name. It also lets the user search emoji by specifying a part-of-speech tagged sense and returns a list of emoji that are tagged with the searched sense. Table~\ref{emojinetstats} lists statistics for EmojiNet. It currently holds a total of 1,074 emoji. It has a total of 3,206 valid sense definitions that are shared among 875 emoji, with an average of 4 senses per emoji. The resource is freely available to the public for research use\footnote{\url{http://emojinet.knoesis.org/}}.

\begin{table}[]
\centering
\caption{EmojiNet Statistics}
\label{emojinetstats}
\begin{tabular}{|l|c|c|c|c|c|c|c|c|}
\hline
\multicolumn{1}{|c|}{\textbf{Emoji Statistic}} & \textbf{\textbf{$u$}} & \textbf{\textbf{$c$}} & \textbf{\textbf{$d$}} & \textbf{\textbf{$K$}} & \textbf{\textbf{$I$}} & \textbf{\textbf{$R$}} & \textbf{\textbf{$H$}} & \textbf{\textbf{$S$}} \\ \hline
    Number of emoji with each feature                        & 1,074         & 845               & 1,074                    &   1,074                & 1,074          & 1,002                        & 705                       & 875                      \\ \hline
                            
    Amount of data stored for each feature             & 1,074         & 845               & 1,074                       & 8,069                & 28,370          & 9,743              &          8                       & 3,206                       \\ \hline
\end{tabular}
\end{table}

\section{Evaluation} \label{sec:eval}
Note that the construction of EmojiNet is based on linking multiple open resources together in an automated fashion. We thus evaluate the automatic creation of EmojiNet. In particular, we evaluate the nearest neighborhood-based image processing algorithm that we used to integrate emoji resources and the most frequent sense-based and most popular sense-based word sense disambiguation algorithms that we used to assign meanings to emoji sense labels. Note that we do not evaluate the usability of EmojiNet based on its performance on a selected task or a benchmark dataset. While sense inventories such as BabelNet have been evaluated on benchmark datasets for WSD or word similarity calculation performance, emoji sense disambiguation and finding emoji similarity are two research problems on their own that have not been explored yet~\cite{miller2016blissfully}. The focus of this paper is not to study or solve those problems. Evaluating the usefulness of EmojiNet should, and will, be addressed once emoji similarity tasks and emoji sense disambiguation tasks are defined with baseline datasets. In lieu of task evaluation, we demonstrate the usefulness of EmojiNet with a use case of how it can be used to address the emoji sense disambiguation problem.

\subsection{Evaluating Image Processing Algorithm}
We next evaluate how well the nearest neighborhood-based image processing algorithm could match each image in $I_{t}$ with the images in $I_{x}$. $I_{x}$ could contain multiple images for a given emoji (7 images per emoji on average), based on different rendering platforms on which the emoji could appear. The set of all different images $I_i \in I_{x}$ that belongs to $e_i$ are tagged with the Unicode representation $u_i$, which is the Unicode representation of $e_i$. For us to find a match between $I_{t}$ and $I_{x}$, we only require one of the multiple images that represents an emoji from $I_{x}$ match with any image from $I_{t}$. Once the matching process is done, we pick the top ranked match based on the dissimilarity of the two matched images and manually evaluate them for equality. While the image processing algorithm we used is naive, it works well in our use case due to several reasons. First, the images of emoji are not complex as they represent a single object (e.g. eggplant \emoji{1F346}) or face (e.g. smiling face \emoji{1F60A}). Second, the emoji images do not contain very complex color combinations as in textures and they are small in size. The image processing algorithm combines color (spectral) information with spatial (position/distribution) information and tends to represent those features well when the images are simple. Third, Euclidean distance ($L2$ distance) prefers many medium disagreements to one large disagreement as in $L1$ distance. Therefore, this nearest neighborhood-based image processing algorithm fits well for our problem.

Manual evaluation of the algorithm revealed that it achieves 98.42\% accuracy in aligning images in $I_{t}$ with $I_{x}$. Out of the 1,074 image instances we checked, our algorithm could correctly find matching images for 1,057 images in $I_{t}$ and it could not find correct matches for 17 images. We checked the 17 incorrect alignments manually and found that eight were clock emoji that express different times of the day. Those images were very similar in color despite the fact that the two arms in the clocks were at different positions. There were three incorrect alignments involving people characters present in the emoji pictures. Those images had minimal differences, which the image processing algorithm could not identify correctly. There were two instances where flags of countries were aligned incorrectly. Again, those flags were very similar in color (e.g. Flag of Russia and Flag of Solvenia). In our error analysis, we identified that the image processing algorithm does not perform correctly when the images are very similar in color but have slight variations in the object(s) it renders. Since the color of the image plays a huge role in this algorithm, the same picture taken in different lighting conditions (i.e. changes in the background color, while the image color stays the same) could decrease the accuracy of the program. However, that does not apply in our case as all the images we considered have a transparent background.

\subsection{Evaluating Word Sense Disambiguation Algorithm}
Here we discuss how we evaluate the most frequent sense-based and most popular sense-based word sense disambiguation algorithms that we used to link Emoji senses with BabelNet sense IDs. We use a manual evaluation approach based on human judges to validate whether a BabelNet sense ID assigned to an emoji sense is valid. We sought the help of two human judges in this task and our judges were graduate students who had either worked on or taken a class on natural language processing. We provided them with all the emoji included in EmojiNet, listing all the valid sense labels extracted from The Emoji Dictionary and their corresponding BabelNet senses (BabelNet sense IDs with definitions) extracted from each WSD approach. The human judges were asked to mark whether they thought that the suggested BabelNet sense ID was the correct sense ID for the emoji sense label listed. If so, they would mark the sense ID prediction as correct, otherwise they would mark it as incorrect. We calculated the agreement between the two judges for this task using Cohen's kappa coefficient\footnote{\url{https://en.wikipedia.org/wiki/Cohen's_kappa}} and obtained an agreement value of 0.7355, which is considered to be a good agreement.

Out of the 3,206 sense labels to disambiguate, the MFS-based method could disambiguate a total of 2,293 sense labels. Our judges analysed these sense labels manually and marked 2,031 of them as correct, with an accuracy of 88.57\% for the MFS-based WSD task. There were 262 cases where the emoji sense was not correctly captured. The correctly dissambiguated sense labels belong to 835 emoji. The 913 sense labels which were not disambiguated in the MFS-based WSD task were considered in a second WSD task, based on the MPS. Our evaluation revealed that the MPS-based WSD task could correctly disambiguate 700 sense labels, with an accuracy of 76.67\%. There were 213 cases where our MPS-based approach failed to correctly disambiguate the sense label. The correctly dissambiguated sense labels belong to 446 emoji.

Table~\ref{wsdstats} integrates the results obtained by both word sense disambiguation algorithms for different part-of-speech tags. The results shows the two WSD approaches we used have performed reasonably well in disambiguating the sense labels in EmojiNet. They have sense-disambiguated with an combined accuracy of 85.18\%. These two methods combined have assigned BabelNet sense IDs to a total of 875 emoji out of the 1,074 emoji we extracted from The Emoji Dictionary website. It shows that our WSD approaches combined have disambiguated senses for 81.47\% of the total number of emoji present in The Emoji Dictionary. However, we do not report on the total number of valid sense labels that we did not extract in our data extraction process since The Emoji Dictionary had 16,392 unique sense labels, which were too big for one to manually evaluate.

\begin{table}[]
\centering
\caption{Word Sense Disambiguation Statistics}
\label{wsdstats}
\begin{tabular}{|l|c|c|c|}
\hline
                                   & \textbf{Correct}         & \textbf{Incorrect}     & \textbf{Total} \\ \hline
Noun                               & 1,271 (83.28\%)            & 255 (16.71\%)            & 1,526          \\ \hline
Verb                               & 735 (84.00\%)            & 140 (16.00\%)           & 875            \\ \hline
Adjective                          & 725 (90.06\%)            & 80 (9.93\%)            & 805            \\ \hline
\multicolumn{1}{|l|}{\textbf{Total}} & \textbf{2,731 (85.18\%)} & \textbf{475 (14.81\%)} & \textbf{3,206} \\ \hline
\end{tabular}
\end{table}

\subsection{EmojiNet at Work} 
We also provide an illustration of EmojiNet in action with a disambiguation of the sense of the \emoji{1F64F} emoji as it is used in two example tweets. We choose this emoji since it is reported as one of the most misused emoji on social media\footnote{\url{http://www.goodhousekeeping.com/life/g3601/surprising-emoji-meanings/}}. The tweets we consider are:
\begin{align*}
T_1&: \text{Pray for my family \emoji{1F64F} God gained an angel today.}  \\
T_2&: \text{Hard to win, but we did it man \emoji{1F64F} Lets celebrate!} 
\end{align*}

EmojiNet lists \texttt{high five(noun)} and \texttt{pray(verb)} as valid senses for the above emoji. For \texttt{high five(noun)}, EmojiNet lists three definitions and for \texttt{pray(verb)}, it lists two definitions. We take all the words that appear in their corresponding definitions as possible context words that can appear when the corresponding sense is being used in a sentence (tweet in this case). For each sense, EmojiNet extracts the following sets of words: 
\begin{align*}
\mathtt{pray(verb)}&: \{worship, thanksgiving, saint, pray, higher, god, confession\} \\
\mathtt{high five(noun)}&: \{palm, high, hand, slide, celebrate, raise, person, head, five\}
\end{align*}

To calculate the sense of the \emoji{1F64F} emoji in each tweet, we calculate the overlap between the words which appear in the tweet with words appearing with each emoji sense listed above. This method is called the Simplified Lesk Algorithm~\cite{vasilescu2004evaluating}. The sense with the highest word overlap is assigned to the emoji at the end of a successful run of the algorithm. We can see that \emoji{1F64F} emoji in $T_1$ will be assigned \texttt{pray(verb)} based on the overlap of words \{god, pray\} with words retrieved from the sense definition of \texttt{pray(verb)} and the same emoji in $T_2$ will be assigned \texttt{high five(noun)} based on the overlap of word \{celebrate\} with words retrieved from the sense definition of \texttt{high five(noun)}. In the above example, we have only shown the minimal set of words that one could extract from EmojiNet. Since we link EmojiNet senses with their corresponding BabelNet senses using BabelNet sense IDs, one could easily utilize other resources available in BabelNet such as related WordNet senses, VerbNet senses, Wikipedia, etc. to collect an improved set of context words for emoji sense disambiguation tasks. It should be emphasized that this example was taken only to show the usefulness of the resource for research directions.

\section{Conclusion and Future Work} \label{sec:con}
This paper presented the construction of EmojiNet, the first ever machine readable sense inventory to understand the meanings of emoji. It integrates four different emoji resources from the Web to extract emoji senses and align those senses with BabelNet. We evaluated the automatic creation of EmojiNet by evaluating (i) the nearest neighborhood-based image processing algorithm used to align different emoji resources and (ii) the most frequent sense-based and the most popular sense-based word sense disambiguation algorithms used to align different emoji senses extracted from the Web with BabelNet. We plan to extend our work in the future by expanding the sense definitions extracted from BabelNet with words extracted from tweets, using a word embedding model trained on tweets that contain emoji. We also plan to evaluate the usability of EmojiNet by first defining the emoji sense disambiguation and emoji similarity finding problems, and then applying EmojiNet to disambiguate emoji senses based on different contexts in which they appear. We are working on applying EmojiNet to improve sentiment analysis and exposing EmojiNet as a web service.

\section*{Acknowledgments}
We are grateful to Sujan Perera for thought-provoking discussions on the topic. We acknowledge partial support from the National Institute on Drug Abuse (NIDA) Grant No. 5R01DA039454-02: ``Trending: Social Media Analysis to Monitor Cannabis and Synthetic Cannabinoid Use'', National Institutes of Health (NIH) award: MH105384-01A1: ``Modeling Social Behavior for Healthcare Utilization in Depression'', and Grant No. 2014-PS-PSN-00006 awarded by the Bureau of Justice Assistance. The Bureau of Justice Assistance is a component of the U.S. Department of Justice's Office of Justice Programs, which also includes the Bureau of Justice Statistics, the National Institute of Justice, the Office of Juvenile Justice and Delinquency Prevention, the Office for Victims of Crime, and the SMART Office. Points of view or opinions in this document are those of the authors and do not necessarily represent the official position or policies of the U.S. Department of Justice, NIH or NIDA.

\bibliographystyle{splncs03}
%\bibliography{SocInfo16}

\end{document}